\documentclass[letterpaper]{article} % DO NOT CHANGE THIS
\usepackage{aaai2026}  % DO NOT CHANGE THIS
\usepackage{times}  % DO NOT CHANGE THIS
\usepackage{helvet}  % DO NOT CHANGE THIS
\usepackage{courier}  % DO NOT CHANGE THIS
\usepackage[hyphens]{url}  % DO NOT CHANGE THIS
\usepackage{graphicx} % DO NOT CHANGE THIS
\usepackage{natbib}  % DO NOT CHANGE THIS AND DO NOT ADD ANY OPTIONS TO IT
\usepackage{caption} % DO NOT CHANGE THIS AND DO NOT ADD ANY OPTIONS TO IT
\usepackage{algorithm}
\usepackage{algorithmic}

\usepackage{newfloat}
\usepackage{listings}
\DeclareCaptionStyle{ruled}{labelfont=normalfont,labelsep=colon,strut=off} % DO NOT CHANGE THIS
\floatstyle{ruled}
\newfloat{listing}{tb}{lst}{}
\floatname{listing}{Listing}
\usepackage{amssymb}
\usepackage{amsmath}
\usepackage{makecell}
\usepackage{multirow}
\usepackage{colortbl}
\definecolor{shadecolor}{gray}{0.9} 

\nocopyright
\title{Pixel-level Quality Assessment for Oriented Object Detection}
\author{
Yunhui Zhu\textsuperscript{\rm 1},
Buliao Huang\textsuperscript{\rm 2}\thanks{Corresponding authors.}
}
\affiliations{
\textsuperscript{\rm 1}School of Computer Science, Nanjing Audit University, Nanjing 211815, China\\
\textsuperscript{\rm 2}School of Computer Science and Engineering, Jinling
Institute of Technology, Nanjing 211169, China\\
yhzhu@nau.edu.cn, hbl1995@jit.edu.cn
}

\usepackage{bibentry}
\begin{document}

\maketitle

\frenchspacing

\begin{abstract}
	Modern oriented object detectors typically predict a set of bounding boxes and select the top-ranked ones based on estimated localization quality. Achieving high detection performance requires that the estimated quality closely aligns with the actual localization accuracy. To this end, existing approaches predict the Intersection over Union (IoU) between the predicted and ground-truth (GT) boxes as a proxy for localization quality. However, box-level IoU prediction suffers from a structural coupling issue: since the predicted box is derived from the detector’s internal estimation of the GT box, the predicted IoU—based on their similarity—can be overestimated for poorly localized boxes. To overcome this limitation, we propose a novel Pixel-level Quality Assessment (PQA) framework, which replaces box-level IoU prediction with the integration of pixel-level spatial consistency. PQA measures the alignment between each pixel’s relative position to the predicted box and its corresponding position to the GT box. By operating at the pixel level, PQA avoids directly comparing the predicted box with the estimated GT box, thereby eliminating the inherent similarity bias in box-level IoU prediction. Furthermore, we introduce a new integration metric that aggregates pixel-level spatial consistency into a unified quality score, yielding a more accurate approximation of the actual localization quality.
	Extensive experiments on HRSC2016 and DOTA demonstrate that PQA can be seamlessly integrated into various oriented object detectors, consistently improving performance (e.g., +5.96\% AP$_{50:95}$ on Rotated RetinaNet and +2.32\% on STD).
\end{abstract}

\section{Introduction}

\label{sec:intro}

\begin{figure}[!t]
	\centering
	\includegraphics[width=0.5\textwidth]{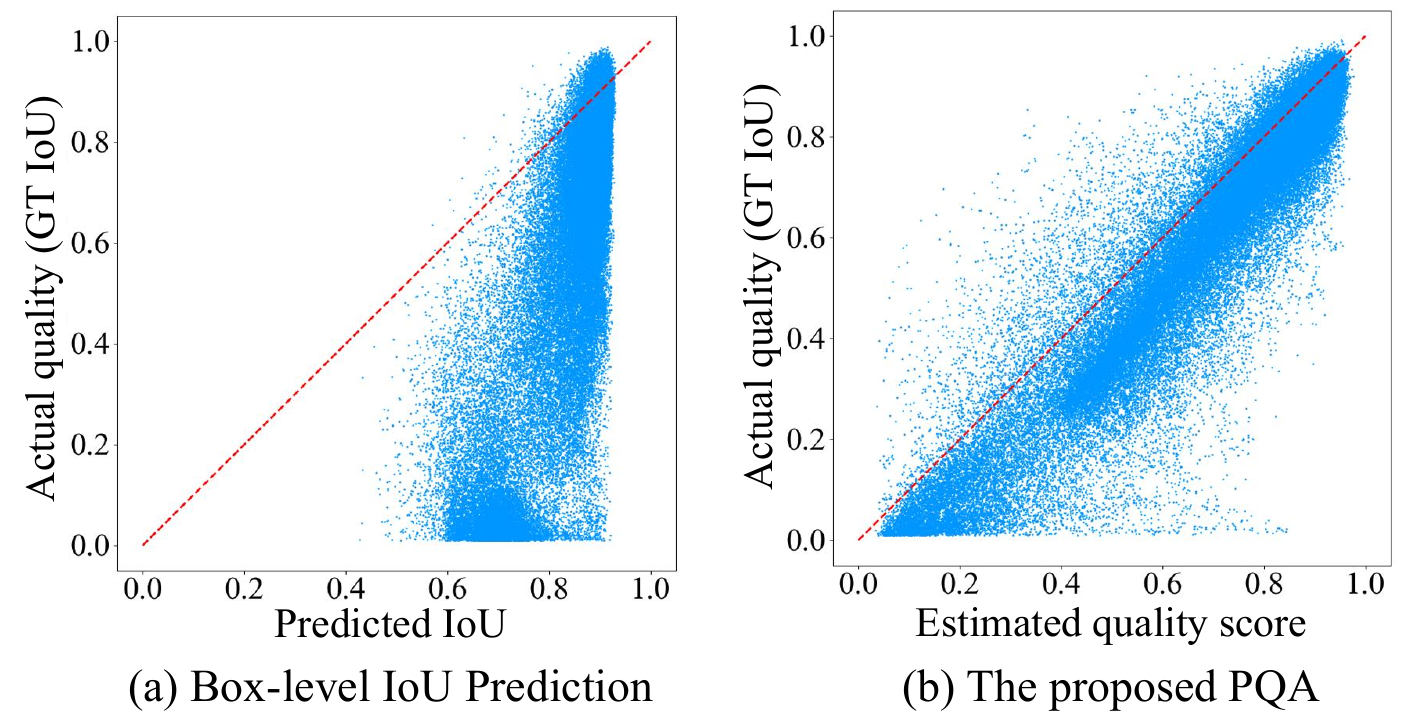}
	\caption{Correlation between the estimated and actual localization quality of predicted oriented boxes.}
	\label{fig:intro_2}
\end{figure}

Oriented object detection, a task focused on accurately classifying and localizing objects with arbitrary orientations in images, has garnered significant attention. Modern oriented object detectors \cite{yu2024spatial, pu2023adaptive, hou2022shape, huang2022general, xie2021oriented} typically generate a set of bounding boxes, rank them according to their classification scores, and select the top-ranked ones as final detections. To achieve high detection performance, it is essential that detections with better localization quality are ranked higher. However, classification scores are primarily optimized for category discrimination and do not necessarily reflect localization accuracy. This discrepancy can lead to suboptimal ranking of predicted boxes, thereby degrading average precision, particularly under high Intersection over Union (IoU) thresholds \cite{kahraman2023correlation}.

Motivated by this issue, a number of studies have aimed to accurately estimate the localization quality. In general, the actual localization quality of a predicted box is quantified by its GT IoU with the GT box. However, since the GT box is unavailable during inference, existing methods \cite{xi2024structure, ming2023task, pu2023rank, zhang2021varifocalnet, li2021generalized} typically predict the IoU between the predicted and GT boxes to approximate the actual localization quality.
Despite its intuitive appeal, this approach may yield unreliable estimates. As illustrated in Figure~\ref{fig:intro_2}(a), predicted boxes with low GT IoU values may still receive high predicted IoU scores. We attribute this to IoU prediction being performed at the box level, where it measures the similarity between the predicted box and the detector’s internal estimation of the GT box.
However, since the predicted box is itself derived from this internal estimation, it naturally shares a high similarity with the estimated GT box. This structural coupling could lead to overestimated IoU scores for poorly localized boxes, undermining the reliability of box-level IoU prediction.

\begin{figure}[!t]
	\centering
	\includegraphics[width=0.49\textwidth]{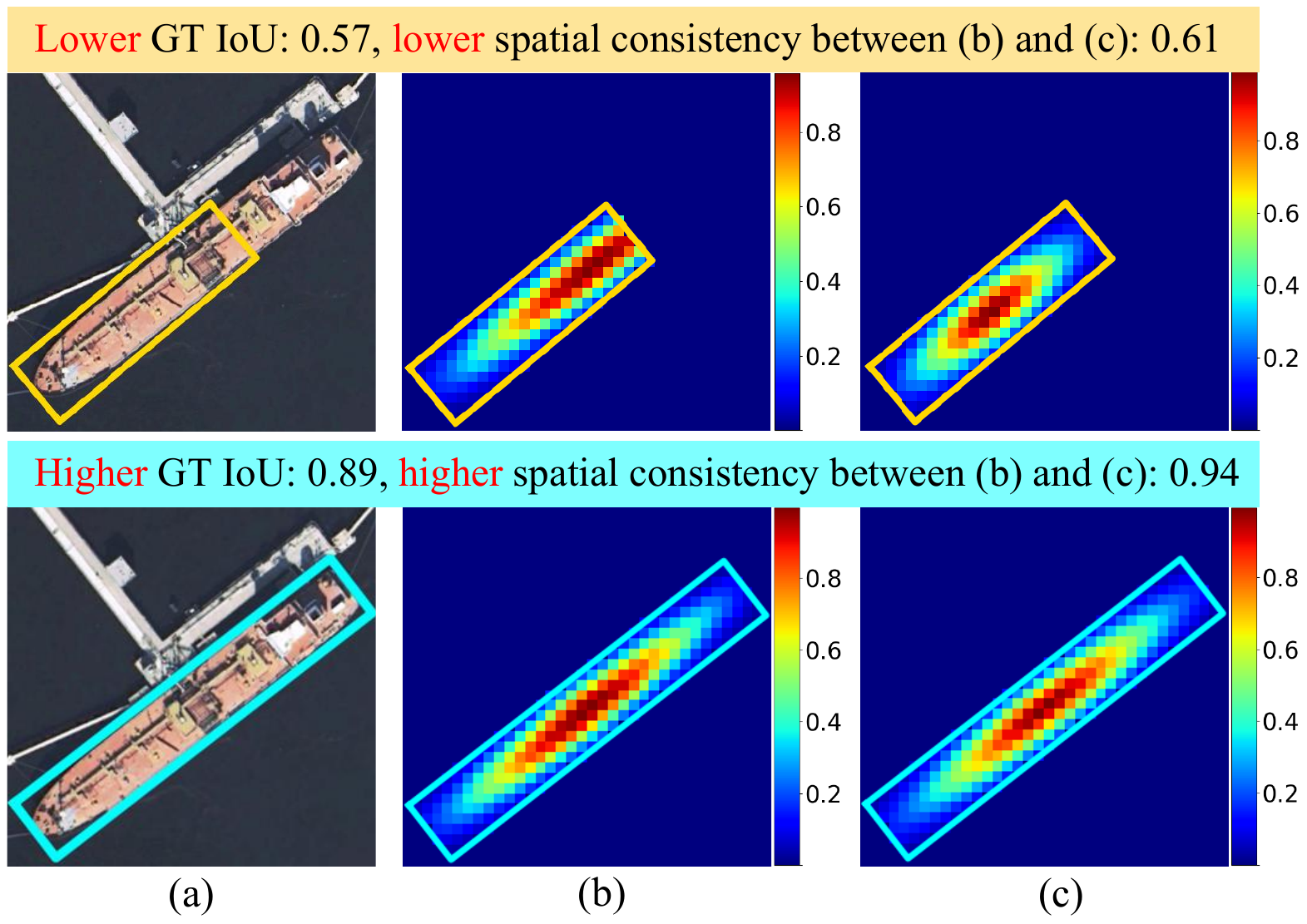}
	\caption{Illustrative examples of how pixel-level spatial consistency correlates with GT IoU. (a) Predicted oriented boxes. (b) Heatmap encoding pixel-wise relative positions to the GT box, where higher values indicate closer proximity to the box center. (c) Heatmap encoding pixel-wise relative positions to the predicted box, represented analogously.
	}
	\label{fig:intro_3}
\end{figure}

To address this limitation, we propose a novel Pixel-level Quality Assessment (PQA) framework, which replaces box-level IoU prediction with the integration of pixel-level spatial consistency. Spatial consistency is measured as the alignment between a pixel’s relative positions to the predicted box and the GT box. This design is motivated by the observation that, for a well-localized predicted box, each pixel’s relative position to the predicted box should closely match its position relative to the GT box. As shown in Figure 2, the predicted box with a higher GT IoU exhibits stronger pixel-level spatial consistency. By operating at the pixel level, PQA avoids directly comparing the predicted box with the estimated GT box, thereby eliminating the inherent similarity bias in box-level IoU prediction and mitigating the resulting overestimation of localization quality. This advantage is further validated in Figure~\ref{fig:intro_2}(b), which shows that the proposed PQA framework establishes a robust positive correlation between the estimated localization quality and the actual GT IoU.
Additionally, to convert pixel-level spatial consistency within a predicted box into a quantifiable quality score, we introduce a novel integration metric that provides a closer approximation to the GT IoU. This aggregation strategy also improves robustness to potential noise in pixel-level spatial consistency, ensuring a more stable estimation of localization quality. This stands in stark contrast to conventional box-level IoU prediction, which directly regresses a scalar IoU value as the quality score, making it inherently sensitive to prediction errors. 

To verify the effectiveness of PQA, we conducted extensive experiments on two popular oriented object detection datasets: HRSC2016 \cite{liu2017high} and DOTA \cite{xia2018dota}. Experiments demonstrate that the proposed PQA can be integrated into various oriented detectors and consistently enhances their performance (e.g., +5.96\% AP$_{50:95}$ on Rotated RetinaNet \cite{lin2017focal} and +2.32\% on STD \cite{yu2024spatial}). 
The main contributions of this paper are summarized as follows:

\begin{itemize}
	
	\item This paper proposes a novel Pixel-level Quality Assessment (PQA) framework that replaces conventional box-level IoU prediction with the integration of pixel-level spatial consistency, effectively eliminating inherent similarity bias and mitigating the overestimation of localization quality.
	
	\item This paper introduces a novel integration metric that aggregates pixel-level spatial consistency into a robust and unified quality score, enabling a more accurate approximation of the GT IoU while remaining resilient to prediction noise.
	
	\item The proposed PQA framework can be seamlessly integrated into existing oriented object detectors, consistently improving localization quality assessment and overall detection performance.	
	
\end{itemize}

\section{Related Work}

\label{sec:related}

\begin{figure*}[!t]
	\centering
	\includegraphics[width=0.95\textwidth]{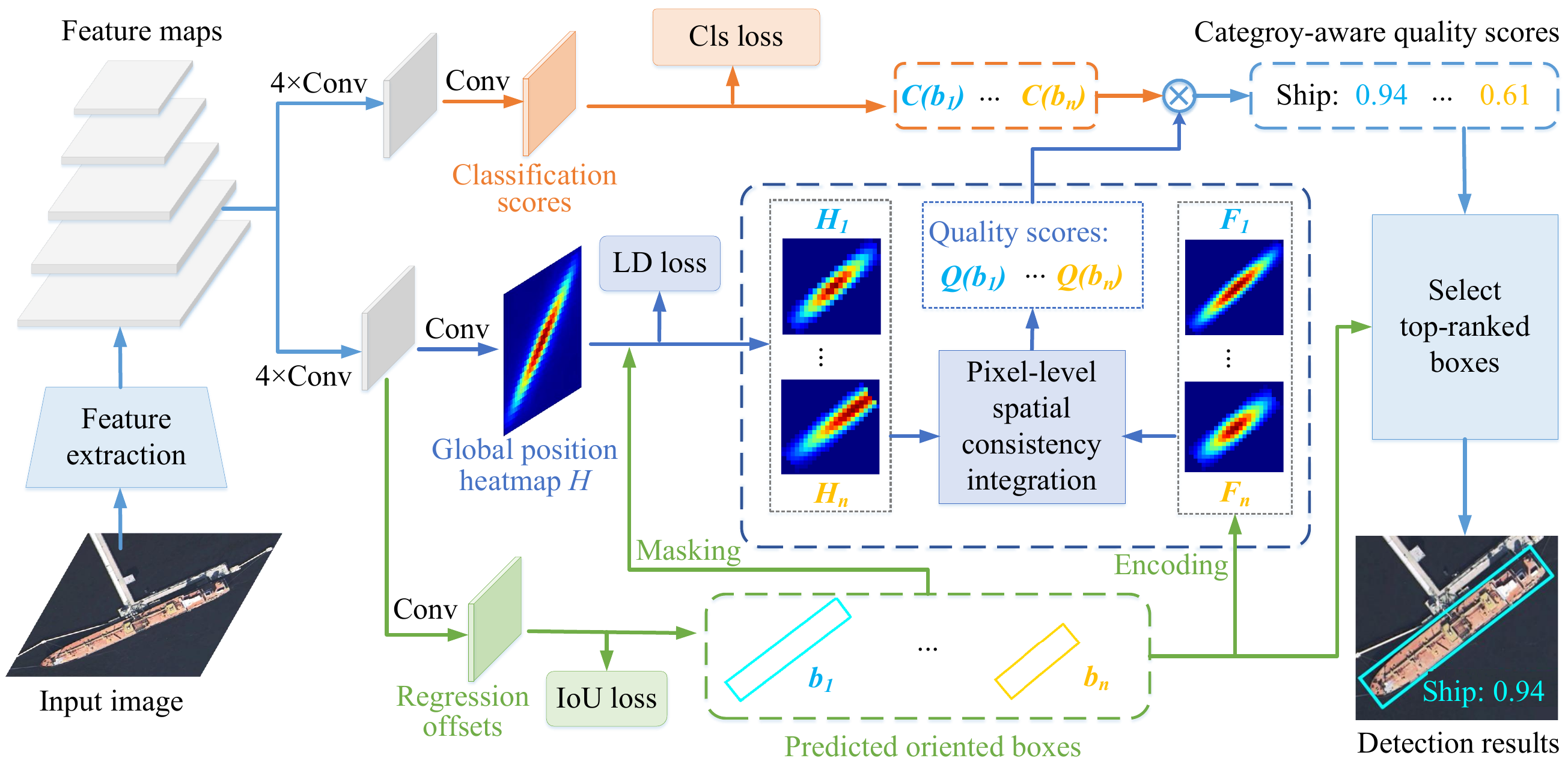}
	\caption{Overall framework of PQA. $H_i$ and $F_i$ provide spatial encodings of pixels' relative positions to the nearest GT box and to the predicted box $b_i$, respectively. }
	\label{fig_pipeline}
\end{figure*}

\subsection{Oriented Object Detection}

Based on generic object detection frameworks, recent studies have sought to enhance oriented detection performance from various perspectives. Some methods have concentrated on effective oriented proposal generation \cite{cheng2022anchor, xie2021oriented, ding2019learning} or more powerful feature extraction \cite{lee2024fred, pu2023adaptive, zhen2023towards, yang2021r3det, han2021align}. 
For example, ARC \cite{pu2023adaptive} proposed an adaptive rotated convolution module to extract high-quality features for objects with varying orientations. In addition, several approaches have explored more flexible representations for oriented objects \cite{yu2023phase, li2022oriented, xu2020gliding}, as well as more effective loss functions for localization \cite{murrugarra2024probabilistic, yang2023kfiou, yang2021learning}.
Moreover, various approaches have investigated advanced label assignment strategies \cite{xu2023dynamic, hou2022shape, huang2022general}. Specifically, SASM \cite{hou2022shape} selected positive training samples based on the shape and distribution information of oriented objects. Furthermore, multiple methods have attempted to enhance the potential of visual transformers \cite{yu2024spatial,zeng2024ars,dai2022ao2}. For example, STD \cite{yu2024spatial} divided the prediction process of oriented boxes into multiple stages to improve the performance of visual transformers.
Despite the progress made by these diverse approaches, most of them relied on the predicted classification scores to assess the quality of predicted oriented boxes and rarely explored the impact of different quality assessment methods on detection performance.

\subsection{Localization Quality Assessment}

Most existing detectors \cite{yu2024spatial, pu2023adaptive, yu2024boundary, hou2022shape, xie2021oriented, lin2017focal} relied solely on the predicted classification scores to evaluate the quality of predicted boxes. However, the inconsistency between classification scores and actual localization quality often impeded the performance of detectors. To address this issue, a number of methods \cite{xi2024structure, li2021generalized, ge2021ota, kim2020probabilistic, tian2019fcos, jiang2018acquisition} introduced an auxiliary branch to predict IoU or centerness values, which were then combined with classification scores. 
Additionally, some methods \cite{ming2023task, zhang2021varifocalnet} jointly modeled classification and IoU prediction, enabling the classification scores to reflect localization quality. For instance, TIOE \cite{ming2023task} introduced a hierarchical alignment label that stratified IoU values into discrete levels to supervise the classifier. Moreover, several works \cite{kahraman2023correlation, oksuz2021rank} designed ranking-based loss functions to enforce the predicted classification scores to follow the same order as the localization quality of the corresponding boxes. 
In contrast to existing approaches, the proposed PQA transforms box-level estimation into the integration of pixel-level spatial consistency, reducing overestimation of localization quality and ensuring a clear positive correlation between the estimated quality score and the GT IoU.

\section{Method}

\label{sec:method}

\subsection{Overview of PQA Framework}

\begin{figure*}[!t]
	\centering
	\includegraphics[width=0.95\textwidth]{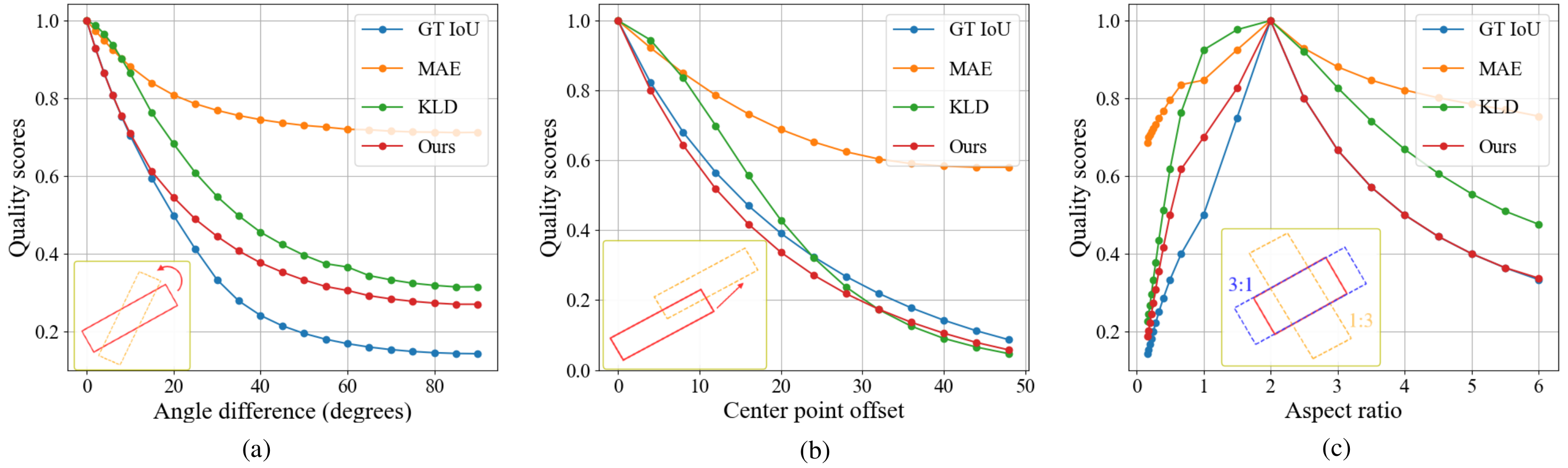}
	\caption{Comparison of quality scores computed using different integration metrics for pixel-level spatial consistency, as the predicted oriented boxes vary in (a) orientation angle, (b) center point offset, and (c) aspect ratio.}
	\label{fig:method_2}
\end{figure*}

The overall framework of PQA is illustrated in Figure~\ref{fig_pipeline}. For each predicted oriented box $b_i$, PQA estimates its localization quality $Q(b_i)$ through the integration of pixel-level spatial consistency within $b_i$.
Specifically, PQA first adds a convolutional layer to predict a global position heatmap $H$, which encodes the relative positions of all pixels with respect to their nearest GT oriented boxes.
Next, for each predicted box $b_i$, PQA applies a spatial mask to $H$ to extract a localized heatmap $H_i$, which activates only the pixels within $b_i$. Simultaneously, the relative positions of these activated pixels to $b_i$ itself are encoded as a heatmap $F_i$.
Finally, PQA introduces an integration metric to aggregate the spatial consistency between $H_i$ and $F_i$, producing the final quality score $Q(b_i)$. This score effectively captures the pixel-level alignment between the predicted box and the GT box, addressing the structural coupling issue inherent in traditional box-level IoU prediction.
The following sections provide detailed descriptions of the prediction of the global position heatmap $H$ and the integration of pixel-level spatial consistency.

\subsection{Global Position Heatmap Prediction}

\label{sec:m1}

As shown in Figure~\ref{fig_pipeline}, PQA incorporates a $3 \times 3$ convolution layer in the localization subnet to predict the global position heatmap $H$ that encodes the relative positions of all pixels with regard to their nearest GT oriented boxes. This subsection details the label definition and loss function (LD loss) used for global position heatmap prediction.

The global position heatmap label is designed to satisfy two conditions: 1) pixels outside the GT oriented box are assigned a value of 0, while those inside are assigned values greater than 0; 2) for pixels inside the GT box, the values vary with the distance to the box center, incorporating both the orientation and size of the GT box to encode the pixel's relative position. Inspired by prior works \cite{yang2021learning, huang2022general} that represent oriented boxes as Gaussian distributions, PQA generates the global position heatmap label based on 2D Gaussian functions, providing a geometry-aware representation.

Formally, given an image $I$ with $M$ objects, let $G=\{g_j\}_{j=1}^M, g_j\in \mathbb{R}^5$ be the set of GT oriented boxes for these objects. For any pixel $p$ in the image $I$, the label $H^*(p)$ of global position heatmap at pixel $p$ is defined as:
{\footnotesize \begin{equation}
		\label{eq:A1}
		H^*(p) = \mathop{\max} \{\Phi\left(p | g_j\right)\}_{j=1}^{M}.
\end{equation}}For any $g_j \in G$, $\Phi\left(p | g_j\right)$ is calculated as follows:
{\footnotesize \begin{equation}
		\label{eq:A2}
		\Phi\left(p | g_j\right) \! = \! 
		\begin{cases}
			e^{-\frac{1}{2} \left(p - g_j^\mu \right)^T \Sigma_{g_j}^{-1} \left(p- g_j^\mu \right)} & \mbox{if $p$ is inside $g_j$}\\
			0 & \mbox{otherwise.}
		\end{cases}
\end{equation}}Here, $g_j^\mu$ denotes the center point of the object $g_j$. $\Sigma_{g_j}$ represents the covariance matrix of the object $g_j$ and is calculated as follows:
{\footnotesize \begin{equation}
		\label{eq:A3}
		\begin{aligned}
			\Sigma_{g_j}^{1/2}& \! = \! U\Lambda U^T \\
			& \! = \! \begin{pmatrix}\cos g_j^\theta \! & \! -\sin g_j^\theta \\ \sin g_j^\theta \! & \! \cos g_j^\theta \end{pmatrix} \!\! \begin{pmatrix} \frac{g_j^w}{4} \! & \! 0 \\ 0 \! & \! \frac{g_j^h}{4} \end{pmatrix} \!\! \begin{pmatrix} \cos g_j^\theta \! & \! \sin g_j^\theta \\ -\sin g_j^\theta \! & \! \cos g_j^\theta \end{pmatrix} \!\! .
		\end{aligned}
\end{equation}}Here, $U$ is the rotation matrix and $\Lambda$ is the eigenvalue matrix. The parameters $g_j^\theta$, $g_j^w$, and $g_j^h$ denote the angle, width, and height of the GT oriented box $g_j$, respectively. 
According to Eqs.~\eqref{eq:A1}-\eqref{eq:A3}, $H^*(p)$ is associated with the relative position of pixel $p$ to the center of the GT box, while accounting for both the box's orientation and size. Notably, beyond the Gaussian-based formulation, $H^*(p)$ can also be defined as the centerness \cite{tian2019fcos} of pixel $p$. In the experimental section, Table~\ref{tab_b3} compares these two label definitions, showing that PQA consistently outperforms existing methods under both settings, which demonstrates the superiority and robustness of the PQA framework.

The LD loss $\mathcal{L}_{ld}$ is inspired by the Focal Loss framework \cite{lin2017focal}, particularly in its treatment of negative samples ($H^*(p) = 0$), where it focuses more on hard negatives while down-weighting the loss from abundant easy negatives. Moreover, to optimize the predicted continuous values for positive samples ($H^*(p) > 0$), we adopt the formulation from GFL \cite{li2020generalized}. Specifically, the LD loss $\mathcal{L}_{ld}$ is defined as:

{\footnotesize \begin{equation}
		\label{eq:loss_ld}
		\begin{aligned}
			\mathcal{L}_{ld} &= \big(1/\textstyle\sum_{p} H^*(p)\big) \textstyle\sum_{p} L\big(H(p), H^*(p)\big), \quad \text{where} \\
			L(x, y) &=
			\begin{cases}
				-\alpha |y - x| \left(y \log x + (1 - y) \log (1 - x)\right), & y > 0\\
				-(1 - \alpha)(1 - x)^\beta \log x, &  y = 0.
			\end{cases}
		\end{aligned}
\end{equation}}Here, $H$ is the predicted global position heatmap. $\alpha$ and $\beta$ are set to 0.25 and 2, following \cite{lin2017focal}.

\subsection{Pixel-level Spatial Consistency Integration}
\label{sec:m2}

This subsection details the integration of pixel-level spatial consistency between $H_i$ and $F_i$ to produce the final quality score $Q(b_i)$. 

Let $B=\{{b_i}\}_{i=1}^{N}, b_i\in \mathbb{R}^5$ be the set of all $N$ predicted oriented boxes. For each predicted box $b_i$, PQA applies a spatial mask to the global position heatmap $H$ to extract a localized heatmap $H_i$, which activates only the pixels within $b_i$. Let $P(b_i)$ represent the set of pixels activated inside $b_i$. The value $H_i(p)$ at any pixel $p$ is defined as follows:
{\footnotesize \begin{equation}
		\label{eq:B2}
		H_i(p) = \begin{cases}
			H(p)  & \mbox{if $p \in P(b_i)$}\\
			0 & \mbox{otherwise.}
		\end{cases}
\end{equation}}Next, PQA encodes the relative positions of these activated pixels to $b_i$ itself into $F_i$ based on Eqs.~\eqref{eq:A2} and \eqref{eq:A3}. $F_i(p)$ is defined as follows:
{\footnotesize \begin{equation}
		\label{eq:B1}
		F_i(p) = \Phi\left(p | b_i\right).
\end{equation}}

Subsequently, the quality score $Q(b_i)$ for the predicted box $b_i$ is computed by aggregating the pixel-level spatial consistency between $H_i$ and $F_i$. Since $Q(b_i)$ is derived from all pixels inside $b_i$, it exhibits robustness to potential noise in pixel-level spatial consistency. In contrast, conventional box-level IoU prediction directly regresses a scalar IoU value as the quality score, making it inherently sensitive to prediction errors. The robustness of our pixel-level integration strategy is further validated in Supplementary Section C. 
Formally, $Q(b_i)$ is computed as follows:
{\footnotesize \begin{equation}
		\label{eq:B3}
		Q(b_i) = S(H_i, F_i),
\end{equation}}which can be instantiated using various integration metrics $S(\cdot)$. Once the quality score $Q(b_i)$ is calculated, it is combined with the classification score $C(b_i)$ of the predicted box to obtain a category-aware quality score:
{\footnotesize\begin{equation}
		CQ(b_i) = C(b_i) \times Q(b_i).
\end{equation}}

The most straightforward integration metric $S(\cdot)$ is the Mean Absolute Error (MAE) between $H_i$ and $F_i$. However, as shown in Figure~\ref{fig:method_2}, the resulting quality score $Q$ is overly smooth and lacks sufficient discriminability, making it highly sensitive to classification noise and thereby degrading detection performance. A detailed analysis is provided in Supplementary Section A.1. To mitigate this issue, we adopt the widely-used Kullback-Leibler Divergence (KLD) as the integration metric. As illustrated in Figure~\ref{fig:method_2}, the KLD-based approach yields more discriminative scores. Nevertheless, KLD inherently emphasizes pixels with higher values in the heatmap, which may not align well with the objectives of our task. A more in-depth discussion is available in Supplementary Section A.2.

To ensure that the calculated quality score aligns more closely with the GT IoU, we have redesigned the integration metric inspired by the conventional IoU calculation used for detection boxes. The proposed metric treats the two heatmaps $H_i$ and $F_i$ as two three-dimensional objects and computes the volume IoU between them, defined as:
{\footnotesize\begin{equation}
		\label{eq:ours}
		S(H_i, F_i) =\frac{\sum_{p\in P(b_i)} \min\left(H_i(p), F_i(p)\right)}{\sum_{p\in P(b_i)} \max\left(H_i(p), F_i(p)\right)}.
\end{equation}}Compared to KLD and MAE, the proposed metric offers better distinction and consistently demonstrates a smaller discrepancy between the calculated quality score and the GT IoU value, as shown in Figure~\ref{fig:method_2}. The experimental section also compares the performance of PQA with different integration metrics.

The total training loss of PQA is defined as follows:
{\footnotesize\begin{equation}
		\label{eq:C4}
		\mathcal{L} = \mathcal{L}_{cls} + \mathcal{L}_{loc} + \lambda\mathcal{L}_{ld},
\end{equation}}
where $\mathcal{L}_{cls}$ and $\mathcal{L}_{loc}$ represent the classification and localization losses, respectively. In practice, $\mathcal{L}_{cls}$ employs Focal loss \cite{lin2017focal}, and $\mathcal{L}_{loc}$ uses Rotated IoU loss \cite{zhou2019iou}. The parameter $\lambda$ serves as a scaling factor, and its various values are evaluated in Supplementary Section C.

\section{Experiments}
\label{sec:experiment}

\subsection{Experimental settings}
{\bf Datasets.}
Two widely used oriented object detection datasets, DOTA-v1.0 \cite{xia2018dota} and HRSC2016 \cite{liu2017high}, are employed for performance evaluation.
DOTA is a large-scale dataset comprising 2806 images and 188,282 instances. It includes fifteen object classes: Plane (PL), Baseball diamond (BD), Bridge (BR), Ground track field (GTF), Small vehicle (SV), Large vehicle (LV), Ship (SH), Tennis court (TC), Basketball court (BC), Storage tank (ST), Soccer-ball field (SBF), Roundabout (RA), Harbor (HA), Swimming pool (SP), and Helicopter (HC). The images in DOTA vary in size from 800$\times$800 to 4000$\times$4000 pixels. In line with common practice, we crop the original images into 1024$\times$1024 patches with an overlap of 200.
HRSC2016 is a widely used dataset for ship detection, comprising 1680 images with dimensions ranging from 339 to 1333 pixels. The dataset includes a training set of 436 images, a validation set of 181 images, and a test set of 444 images. 
Following common practice for the two datasets, both the training and validation sets are used for training, while the test set is reserved for testing. 
For performance evaluation, we report the average precision results under IoU thresholds of 0.5 (AP$_{50}$) and 0.75 (AP$_{75}$), as well as the average precision averaged over IoU thresholds from 0.5 to 0.95 in steps of 0.05 (AP$_{50:95}$).

{\bf Implementation Details.}
The implementation of PQA is based on the MMRotate toolbox  \cite{zhou2022mmrotate}. The initial learning rate is set to $2.5 \times 10^{-3}$ with two images per mini-batch. For the DOTA-v1.0 dataset, the training runs for 12 epochs, with the learning rate reduced by a factor of 10 after 8 and 11 epochs. For the HRSC2016 dataset, training lasts for 72 epochs, with the learning rate decaying after 48 and 66 epochs. During training, only random horizontal, vertical, and diagonal flipping are used for image augmentation, with no additional tricks employed. No data augmentation is applied during testing. Unless otherwise specified, the ablation study is conducted on the DOTA-v1.0 dataset, and results are reported on the validation set. The baseline model is Rotated RetinaNet \cite{lin2017focal} with a ResNet-50 backbone \cite{he2016deep}. It adopts the Rotated IoU loss \cite{zhou2019iou} instead of the Smooth L$_1$ loss \cite{girshick2015fast} for localization optimization, and employs SimOTA \cite{ge2021ota} in place of the Max-IoU strategy \cite{ren2015faster} for label assignment. Classification scores are directly used as estimates of localization quality.

\subsection{Ablation studies}
\begin{table}[!t]
	\scriptsize
	\centering
	\renewcommand{\arraystretch}{1.0}
	\setlength{\tabcolsep}{0.8mm}{
		\begin{tabular}{c|ccc|ccc}
			\hline
			\multirow{2}{*}{Method} & \multicolumn{3}{c|}{DOTA-v1.0} & \multicolumn{3}{c}{HRSC2016}  \\
			\cline{2-7}
			& AP$_{50:95}$ & AP$_{50}$ & AP$_{75}$ & AP$_{50:95}$ & AP$_{50}$ & AP$_{75}$ \\
			\hline
			\hline
			Baseline & 36.07 & 65.17 & 34.00    & 62.48 & 90.09 & 77.21 \\
			\hline
			{\bf Auxiliary Head} &  &  &  &   &  &  \\
			Centerness & 36.77 & 66.42 & 35.27    & 62.92 & 89.83 & 76.71  \\
			IoU & 36.79 & 66.52 & 35.12    & 63.16 & 89.77 & 75.90  \\
			SOOD & 36.53 & 66.21 & 34.95    & 63.19 & 89.90 & 76.82  \\
			\hline
			{\bf IoU-aware Cls.}  &  &  &  &   &  &  \\
			TIOE & 36.03 & 65.74 & 33.88     & 63.98 & 90.12 & 76.71  \\
			VFL \cite{zhang2021varifocalnet} &  36.54 & 65.97  &  34.03    &  62.08 &  89.69 & 76.54  \\
			\hline
			{\bf Ranking-based Loss}  &  &  &  &   &  &  \\
			Correlation Loss & 36.06  & 64.79  & 34.81     & 63.61  & 89.84  & 77.35  \\
			RS Loss & 35.91  & 65.77  & 33.62 &     63.05 & 90.03  & 76.16  \\
			\hline
			\rowcolor{shadecolor} 
			PQA & {\bf 38.46} $_{(\uparrow 2.39)}$ & {\bf 67.43} & {\bf 36.44}     & {\bf 66.28} $_{(\uparrow 3.80)}$ & {\bf 90.17} & {\bf 79.07} \\
			\hline
		\end{tabular}
	}
	\caption{Comparison with existing quality assessment methods.}
	\label{tab_b1}
\end{table}

\begin{table}[!t]
	\scriptsize
	\centering
	\renewcommand{\arraystretch}{1.0}
	\setlength{\tabcolsep}{0.8mm}{
		\begin{tabular}{c|cc|ccc|ccc}
			\hline
			\multirow{2}{*}{Method} & \multicolumn{2}{c|}{Label Definition} & \multicolumn{3}{c|}{Integration Metric} & \multirow{2}{*}{AP$_{50:95}$} & \multirow{2}{*}{AP$_{50}$} & \multirow{2}{*}{AP$_{75}$} \\
			\cline{2-6}
			& Centerness & Gaussian & MAE & KLD & Ours &  &  & \\ 
			\hline
			\hline
			Baseline &  &  &  &  &  & 36.07 & 65.17 & 34.00 \\
			\hline
			\multirow{4}{*}{PQA} &  & \checkmark &  \checkmark &  &  &  37.97 $_{{(\uparrow 1.90)}}$ & 67.22 & 36.10 \\
			& & \checkmark &  & \checkmark &  & 38.04 $_{{(\uparrow 1.97)}}$ & 67.16 & 36.18 \\
			& \checkmark &  &  &  & \checkmark & 38.08 $_{{(\uparrow 2.01)}}$ & 67.41 & 35.78  \\
			& & \checkmark &  &  & \checkmark & {\bf 38.46} $_{{(\uparrow 2.39)}}$ & {\bf 67.43} & {\bf 36.44} \\
			\hline
			
	\end{tabular}}
	\caption{Performance of PQA with different configurations.}
	\label{tab_b3}
\end{table}

\begin{table}[!t]
	\scriptsize
	\centering
	\renewcommand{\arraystretch}{1.0}
	\setlength{\tabcolsep}{0.6mm}{
		\begin{tabular}{c|c|ccc|ccc}
			\hline
			\multirow{2}{*}{Method} & \multirow{2}{*}{PQA} & \multicolumn{3}{c|}{DOTA-v1.0} & \multicolumn{3}{c}{HRSC2016} \\
			\cline{3-8}
			& & AP$_{50:95}$ & AP$_{50}$ & AP$_{75}$ & AP$_{50:95}$ & AP$_{50}$ & AP$_{75}$ \\
			\hline
			\hline
			SASM &  &  33.61  & 60.40 & 32.71       & 59.89 & 86.96 & 68.50 \\
			\rowcolor{shadecolor} 
			SASM & {\bf \checkmark} & {\bf 35.49} $_{{(\uparrow 1.88)}}$ & {\bf 62.38} & {\bf 34.93}      & {\bf 66.61} $_{{(\uparrow 6.72)}}$ & {\bf 89.99} & {\bf 79.49} \\
			\hline
			Rotated RetinaNet &   & 35.82 & 64.76 & 33.79    & 50.33 & 85.17 & 49.61 \\
			\rowcolor{shadecolor} 
			Rotated RetinaNet & {\bf \checkmark}   & {\bf 37.28}  $_{{(\uparrow 1.46)}}$ & {\bf 65.82} & {\bf 35.81}         & {\bf 56.29} $_{{(\uparrow 5.96)}}$ & {\bf 87.85} & {\bf 62.34}  \\
			\hline
			Rotated FCOS &   & 37.09 & 67.44 & 35.16      & 60.78 & 89.11 & 74.08  \\
			\rowcolor{shadecolor} 
			Rotated FCOS & {\bf \checkmark} & {\bf 38.25} $_{{(\uparrow 1.16)}}$ & {\bf 69.03} & {\bf 35.17}    & {\bf 65.49} $_{{(\uparrow 4.71)}}$ & {\bf 90.14} & {\bf 79.48}  \\
			\hline
			STD &   & 45.69 & 74.87 & 48.06        & 70.07 & 90.49 & 88.69  \\
			\rowcolor{shadecolor} 
			STD & {\bf \checkmark} & {\bf 46.82} $_{{(\uparrow 1.13)}}$ & {\bf 74.99} & {\bf 49.97}        & {\bf 72.39} $_{{(\uparrow 2.32)}}$ & {\bf 90.63} & {\bf 89.41} \\
			\hline
	\end{tabular}}
	\caption{Performance of PQA integrated into various oriented object detectors.}
	\label{tab_b6}
\end{table}

\begin{table}[t]
	\scriptsize
	\centering
	\renewcommand{\arraystretch}{1.0}
	\setlength{\tabcolsep}{1.0mm}{
		\begin{tabular}{c|ccc|ccc}
			\hline
			Method & AP$_{50:95}$ & AP$_{50}$ & AP$_{75}$ & Params (M) $\downarrow$ & FLOPs (G) $\downarrow$ & FPS $\uparrow$ \\
			\hline
			\hline
			Baseline & 36.07 & 65.17 & 34.00 & 36.05 & 207.87 & 54.2 \\
			PQA      & 38.46 $_{{(\uparrow 2.39)}}$ & 67.43 & 36.44 & 36.07 & 208.28  & 46.9 \\
			PQA-Lite & 37.69 $_{{(\uparrow 1.62)}}$ & 65.27 & 36.48 & 36.07 & 208.28  & 52.0 \\
			\hline
	\end{tabular}}
	\caption{Comparison of full and lightweight PQA variants.}
	\label{tab:lite}
\end{table}

\begin{table*}[!t]
	\scriptsize
	\centering
	\renewcommand{\arraystretch}{1.0}
	\setlength{\tabcolsep}{0.7mm}{
		\begin{tabular}{c|c|ccccccccccccccc|ccc}
			\hline
			Method & Backbone & PL & BD & BR & GTF & SV & LV & SH & TC & BC & ST & SBF & RA & HA & SP & HC & AP$_{50:95}$ & AP$_{50}$ & AP$_{75}$ \\
			\hline
			\hline
			Rotated FCOS \cite{tian2019fcos} & R-50 & 89.06 & 76.97 & 47.92 & 58.55 & 79.78 & 76.95 & 86.90 & 90.90  & 84.87 & 84.58 & 57.11 & 64.68 & 63.69 & 69.38 & 46.87 & 39.80 & 71.88 & 37.30  \\
			Gliding Vertex \cite{xu2020gliding} & R-50 & 89.20 & 75.92 & 51.31 & 69.56 & 78.11 & 75.63 & 86.87 & 90.90  & 85.40 & 84.77 & 53.36 & 66.65 & 66.31 & 69.99 & 54.39  & 39.52 & 73.22 & 37.47 \\
			S$^2$A-Net \cite{han2021align} & R-50 & 89.25 & 81.19 & 51.55 & 71.39 & 78.61 & 77.37 & 86.77 & 90.89  & 86.28 & 84.64 & 61.21 & 65.65 & 66.07 & 67.57 & 50.18 & 39.05 & 73.91 & 35.52  \\
			R$^3$Det  \cite{yang2021r3det} & R-50 & 89.29 & 75.21 & 45.41 & 69.23 & 75.53 & 72.89 & 79.28 & 90.88 & 81.02 & 83.25 & 58.81 & 63.15 & 63.40 & 62.21 & 37.41 & 37.82 & 69.80 & 36.59  \\
			SASM \cite{hou2022shape} & R-50 & 87.51 & 80.15 & 51.07 & 70.35 & 74.95 & 75.80 & 84.23 & 90.90 & 80.87 & 84.93 & 58.51 & 65.59 & 69.74 & 70.18 & 42.31 & 43.01 & 72.47 & 44.21  \\
			O-Reppoints \cite{li2022oriented} & R-50 & 87.78 & 77.67 & 49.54 & 66.46 & 78.51 & 73.11 & 86.58 & 90.86 & 83.75 & 84.34 & 53.14 & 65.63 & 63.70 & 68.71 & 45.91 & 40.88 & 71.71 & 41.39  \\
			H2RBox \cite{yangh2rbox} & R-50 & 88.16 & 80.47 & 40.88 & 61.27 & 79.78 & 75.25 & 84.40 & 90.89 & 80.05 & 85.35 & 58.91 & 68.46 & 63.67 & 71.87 & 47.18 & 41.49 & 71.77 & 41.42  \\
			KFIoU \cite{yang2023kfiou} & R-50 & 89.20 & 76.40 & 51.64 & 70.15 & 78.31 & 76.43 & 87.10 & 90.88 & 81.68 & 82.22 & {\bf 64.65} & 64.84 & 66.77 & 70.68 & 49.52 & 41.70 & 73.37 & 42.71  \\
			AO2-DETR \cite{dai2022ao2} & R-50 & 87.99 & 79.46 & 45.74 & 66.64 & 78.90 & 73.90 & 73.30 & 90.40 & 80.55 & 85.89 & 55.19 & 63.62 & 51.83 & 70.15 & 60.04 & 33.31 & 70.91 & 22.60  \\
			RoI Trans. \cite{ding2019learning} & Swin-T & 88.44 & {\bf 85.53} & 54.56 & 74.55 & 73.43 & 78.39 & 87.64 & 90.88 & {\bf 87.23}  & {\bf 87.11} & 64.25 & 63.27 & {\bf 77.93} & {\bf 74.10} & 60.03 & 47.60 & 76.49 & 50.15  \\
			PSC \cite{yu2023phase} & R-50 & {\bf 89.65} & 83.80 & 43.64 & 70.98 & 79.00 & 71.35 & 85.08 & 90.90 & 84.28 & 82.51 & 60.64 & 65.06 & 62.52 & 69.61 & 54.00 & 43.98 & 72.87 & 46.18  \\
			ARS-DETR \cite{zeng2024ars} & Swin-T & 87.65 & 76.54 & 50.64 & 69.85 & 79.76 & 83.91 & 87.92 & 90.26 & 86.24 & 85.09 & 54.58 & 67.01 & 75.62 & 73.66 & 63.39 & 47.77 & 75.47 & 51.77  \\
			
			Oriented RCNN \cite{xie2021oriented} & R-50 & 89.47 & 83.84 & 55.42 & 72.93 & 78.71 & 83.86 & 88.00 & 90.90 & 87.07 & 84.45 & 63.26 & 69.07 & 76.21 & 69.03 & 51.17 & 46.88 & 75.67 & 50.41  \\
			
			Oriented RCNN \cite{pu2023adaptive} & ARC-R50  & 89.46 & 82.84 & 54.64 & {\bf 75.57} & 79.24 & 84.25 & 88.16 & 90.90 & 86.21 & 85.69 & 63.80 & 65.92 & 74.70 & 69.94 & 60.80 & 47.63 & 76.81 & 51.72 \\
			
			\hline
			STD \cite{yu2024spatial} & ViT-B & 89.56 & 83.18 & {\bf 56.45} & 70.45 & 79.39 &{\bf 85.84} & {\bf 88.39} & 90.88 & 85.32 & 86.54 & 61.56 & 68.29 & 76.81 & 72.92 & 69.81 & 48.26 & 77.68 & 53.12 \\
			\rowcolor{shadecolor} 
			STD + PQA & ViT-B & 89.54 & 82.58 & 55.55 & 72.89 &{\bf 80.48} &  85.63 & 88.28 & {\bf 90.91} & 86.01 & 85.16 & 59.91 & {\bf 70.41} & 77.79 & 71.81 & {\bf 73.77} & {\bf 49.47} $_{{(\uparrow 1.21)}}$ & {\bf 78.05} & {\bf 53.91} \\
			\hline
			
	\end{tabular}}
	\caption{
		Performance comparison on the DOTA-v1.0 dataset. R-50, Swin-T, and ViT-B represent ResNet50 \cite{he2016deep}, Swin-Tansformer \cite{liu2021swin}, Vision Transformer-Base \cite{dosovitskiy2020image} respectively.
	}
	\label{table:dota}
\end{table*}

\begin{figure*}[!t]
	\centering
	\includegraphics[width=0.95\textwidth]{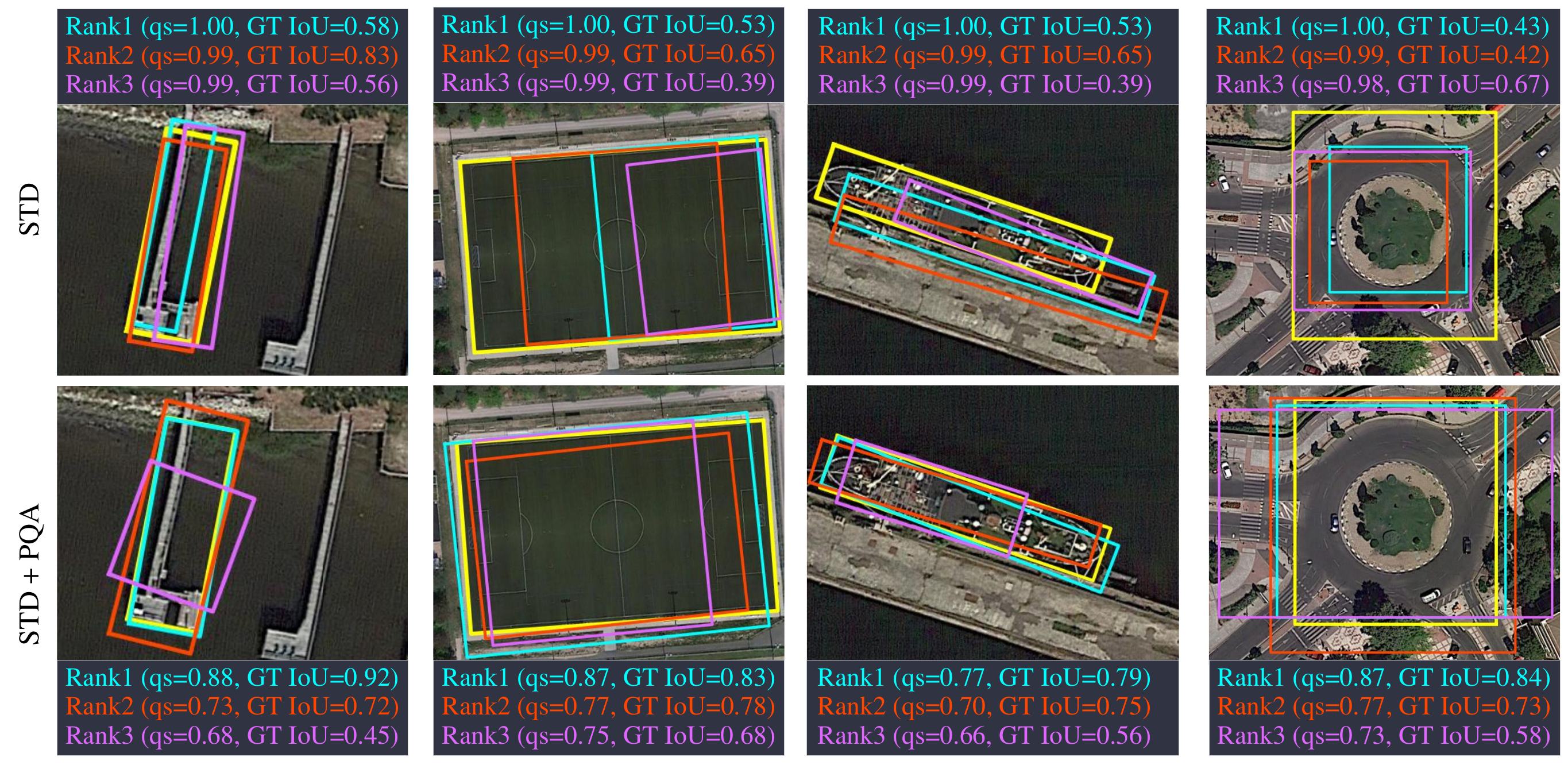} 
	\caption{Visualization of top-ranked predicted oriented boxes with their estimated quality scores (qs) and actual localization quality (GT IoU) before (row 1) and after (row 2) integrating PQA into STD. Predicted boxes are ranked based on qs. Yellow denotes GT boxes; cyan, red, and purple indicate the rank-1, rank-2, and rank-3 predicted boxes, respectively.
	}
	\label{fig:vis}
\end{figure*}

{\bf Compare with Other Quality Assessment Methods.} 
Existing quality assessment methods can be broadly categorized into three types: 1) Auxiliary head methods that introduce an additional branch to predict IoU or centerness values; 2) IoU-aware classification methods that use GT IoU as the classification confidence label for positive samples; and 3) ranking-based loss methods that enforce consistency between the rankings of predicted classification scores and GT IoUs. 
To validate the effectiveness of the proposed PQA framework, we compare it with recent and representative works from each category, including Centerness \cite{tian2019fcos}, IoU \cite{jiang2018acquisition}, SOOD \cite{xi2024structure}, TIOE \cite{ming2023task}, VFL \cite{zhang2021varifocalnet}, Correlation Loss \cite{kahraman2023correlation}, and RS Loss \cite{oksuz2021rank}.
For fair comparison, all methods share the same detection architecture as PQA, differing only in their quality assessment strategies. As shown in Table~\ref{tab_b1}, PQA achieves significant improvements of 2.39\% and 3.80\% in AP$_{50:95}$ over the baseline on the DOTA and HRSC2016 datasets, respectively. Furthermore, it outperforms all compared methods by at least 1.67\% and 2.30\% on the two datasets, demonstrating the superiority and effectiveness of PQA.

{\bf Different Configurations of PQA.}
We evaluate different label definitions for the global position heatmap and various integration metrics for pixel-level spatial consistency. Both centerness \cite{tian2019fcos} and the Gaussian-based formulation can be employed as labels for the global position heatmap, while commonly used metrics such as MAE and KLD, together with our proposed metric (Eq.~\eqref{eq:ours}), serve as integration metrics.
The comparison results are presented in Table~\ref{tab_b3}. It can be observed that, regardless of the choice of label definition or integration metric, PQA consistently outperforms the baseline by more than 1.9\%, demonstrating its robustness and overall superiority. Moreover, our proposed integration metric achieves the best performance, which is consistent with the observation in Figure~\ref{fig:method_2}.

{\bf Compatibility for Various Oriented Detectors.} To verify the compatibility of PQA, it is integrated into various oriented detectors: SASM\cite{hou2022shape}, Rotated RetiaNet \cite{lin2017focal}, Rotated FCOS \cite{tian2019fcos}, and STD \cite{yu2024spatial}, encompassing both anchor-based and anchor-free, single-stage and two-stage methods. Table~\ref{tab_b6} presents the numerical results of these detectors with and without PQA. The results show that PQA consistently brings significant performance improvements across various oriented detectors, particularly in AP$_{75}$ and AP$_{50:95}$.

{\bf Computational Overhead and Lightweight Variant.}
While PQA yields notable gains in AP, it incurs a modest computational overhead. As presented in Table~\ref{tab:lite}, it introduces only 0.02M additional parameters and 0.41G FLOPs compared to the baseline, with inference speed decreasing from 54.2 to 46.9 FPS. This drop is primarily due to the integration of pixel-level spatial consistency across all predicted boxes and their internal pixels. To alleviate this overhead, we propose a lightweight variant, PQA-Lite, which restricts spatial consistency integration to high-confidence predicted boxes and reduces the number of sampled pixels per box. PQA-Lite achieves performance comparable to the full version while improving inference speed to 52.0 FPS, highlighting the efficiency and adaptability of PQA. Further details of PQA-Lite are provided in Supplementary Section~B.

\subsection{Comparison with state-of-the-arts}

\begin{table}[!t]
	\scriptsize
	\centering
	\renewcommand{\arraystretch}{1.0}
	\setlength{\tabcolsep}{0.8mm}{
		\begin{tabular}{c|c|ccc}
			\hline
			Method & Backbone & AP$_{50:95}$ & AP$_{50}$ & AP$_{75}$ \\
			\hline 
			\hline
			Rotated RetinaNet \cite{lin2017focal} & R-50 & 50.33 & 85.17 & 49.61 \\
			Rotated FCOS \cite{tian2019fcos} & R-50 & 60.78 & 87.85 & 62.34 \\  
			S$^2$A-Net \cite{han2021align} & R-50 & 57.92 & 90.18 & 67.11  \\	
			SASM \cite{hou2022shape} & R-50 & 59.89 & 86.96 & 68.50 \\
			AO2-DETR \cite{dai2022ao2} & R-50 & 51.13 & 87.70 & 54.90  \\
			PSC \cite{yu2023phase} & R-50 & 67.57 & 90.06 & 78.56  \\
			ProbIoU \cite{murrugarra2024probabilistic} & R-50 & 61.65  & 90.23 & 78.46 \\
			Oriented RCNN \cite{xie2021oriented} & R-50  & 70.18 & 90.33 & 89.09 \\
			Oriented RCNN \cite{pu2023adaptive} & ARC-R50 & 71.10 & 90.47 & 89.32 \\
			\hline
			STD \cite{yu2024spatial} & ViT-B & 70.07 & 90.49 & 88.69  \\
			\rowcolor{shadecolor} 
			STD + PQA & ViT-B & {\bf 72.39} $_{{(\uparrow 2.32)}}$ & {\bf 90.63} & {\bf 89.41} \\
			\hline
			
	\end{tabular}}
	\caption{Performance comparison on HRSC2016.}
	\label{table:hrsc}
\end{table}

{\bf DOTA-v1.0.}
Table~\ref{table:dota} provides a comparison between PQA and state-of-the-art methods on the DOTA-v1.0 dataset. To ensure a fair evaluation, all methods adopt the same data augmentation strategies and employ single-scale training and testing. When integrated into the advanced detector STD, PQA yields a 1.21\% improvement in AP$_{50:95}$, achieving a competitive AP$_{50}$ of 78.05\% and AP$_{50:95}$ of 49.47\% without relying on any test-time enhancements.

{\bf HRSC2016.}
Table~\ref{table:hrsc} presents the evaluation results on the HRSC2016 dataset. Integrating PQA into the detector STD consistently improves performance, yielding a 2.32\% gain in AP$_{50:95}$ and achieving a remarkable AP$_{50}$ of 90.63\% and AP$_{50:95}$ of 72.39\%. These results are highly competitive and compare favorably with existing approaches.

\subsection{Visualization}

Figure~\ref{fig:vis} compares the ranking of predicted oriented boxes before (row 1) and after (row 2) integrating PQA into STD. During inference, box ranking is determined by the estimated quality score (denoted as qs in Figure~\ref{fig:vis}), and a reliable detector should yield consistent rankings with the GT IoU values. Prior to applying PQA, the qs in STD is derived from the classification score, which shows limited variation (e.g., 1.0 vs. 0.99) and insufficient discriminability. As shown in row 1, some rank-1 boxes still have lower GT IoU values, indicating that classification-based qs fails to reflect actual localization quality. In contrast, after incorporating PQA (row 2), the estimated qs closely aligns with the GT IoU, leading to more accurate and consistent rankings. This highlights the effectiveness of PQA in enhancing localization quality assessment.

\section{Conclusion}
\label{sec:conclusion}
This paper presents a Pixel-level Quality Assessment (PQA) framework for oriented object detection. PQA replaces conventional box-level IoU prediction with the integration of pixel-level spatial consistency. By operating at the pixel level, PQA effectively eliminates the structural coupling issue and ensures a robust positive correlation between the estimated and actual localization quality. In addition, a novel integration metric is proposed to convert pixel-level spatial consistency into a unified quality score, offering a closer approximation to the GT IoU. Experiments on benchmark datasets demonstrate that PQA can be seamlessly integrated into various oriented object detectors and consistently improve their performance, highlighting the effectiveness and generalizability of the proposed approach.

\bibliography{ref}

\clearpage

\twocolumn[
\begin{center}
	\vspace{1em}
	{\LARGE \bfseries Supplementary Material\par}
	\vspace{2em}
\end{center}
]

In the supplementary material, we first analyze the impact of different integration metrics for pixel-level spatial consistency, including Mean Absolute Error (MAE) and Kullback-Leibler Divergence (KLD). Next, we introduce a lightweight variant of our method, PQA-Lite, and provide detailed ablation studies on its key configurations. Finally, we conduct a comprehensive robustness analysis under controlled perturbations and examine the influence of the balancing parameter $\lambda$ in training.

\section{A. Discussion on Integration Metrics}

In this section, we examine two alternative integration metrics—MAE and KLD—and analyze their limitations in the context of pixel-level quality assessment. Through illustrative examples, we show that although both metrics are conceptually straightforward, they exhibit undesirable behaviors such as oversmoothing or biased sensitivity to spatial discrepancies. These observations highlight the need for a more suitable metric specifically tailored to the requirements of our task.

\subsection{A.1. Discussion on MAE-based integration metric}

The most straightforward integration metric $S(\cdot)$ is based on the MAE between heatmaps $H_i$ and $F_i$. The formulation of MAE between corresponding pixels in heatmaps $H_i$ and $F_i$ can be expressed as:
\begin{equation}
	S(H_i, F_i) = 1 - \frac{1}{|P(b_i)|}\sum_{p \in P(b_i)} |H_i(p)- F_i(p)|,
\end{equation} where $P(b_i)$ represent the set of pixels activated inside predicted box $b_i$, $H_i(p)$ and $F_i(p)$ refer to the heatmap value of pixel $p$ in heatmaps $H_i$ and $F_i$, respectively.

However, as shown in Figure 4 of the main manuscript, the calculated quality score tends to be overly smooth and lacks sufficient distinction. Since the calculated quality score is multiplied by the classification score to obtain the category-aware quality score, this lack of distinction makes the quality score sensitive to external noise, which can adversely impact the results, particularly when the classification score is inaccurate. 

\begin{figure}[t]
	\centering
	\includegraphics[width=0.98\columnwidth]{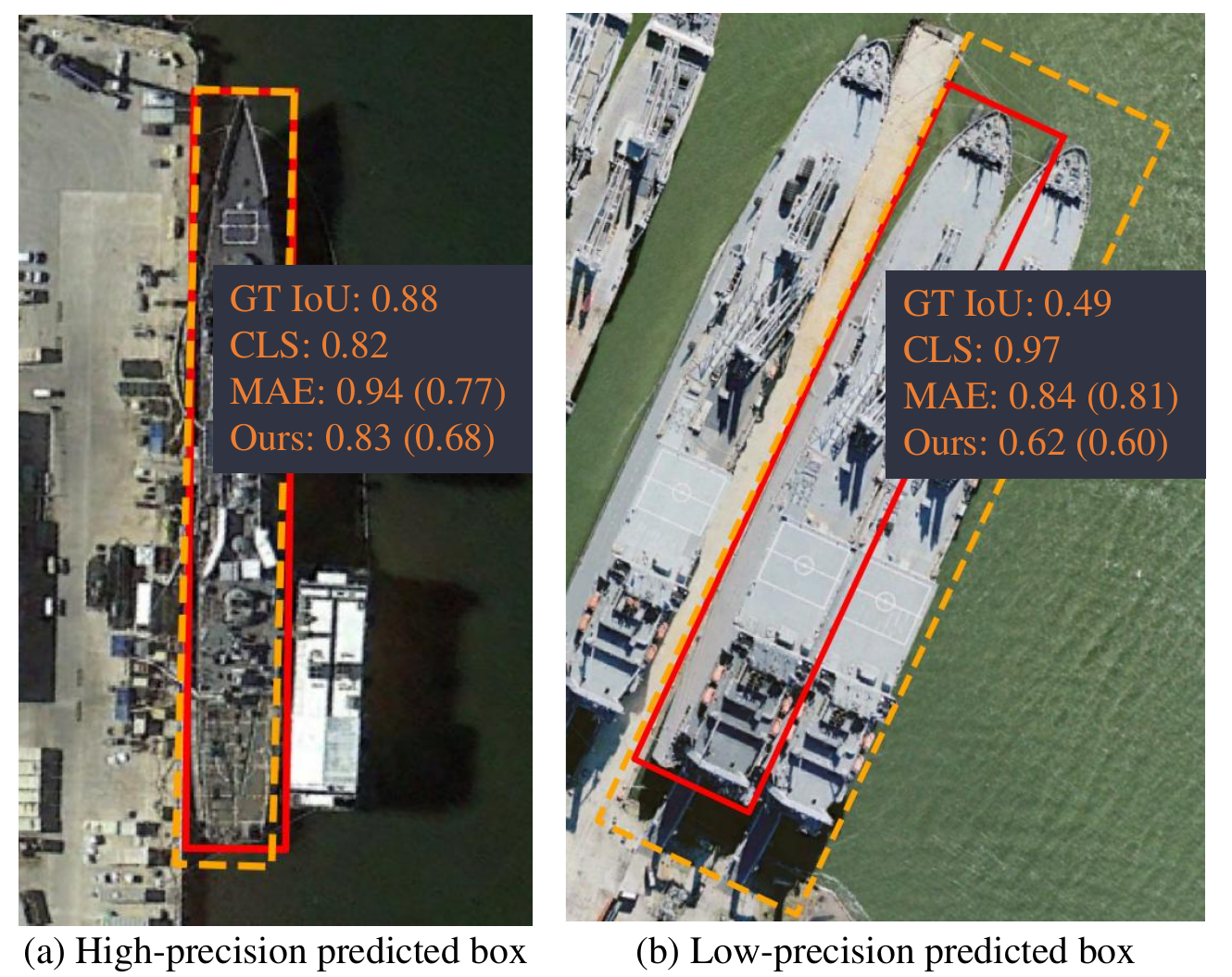}
	\caption{Illustrative case for the influence of inaccurate classification scores on MAE. ``GT IoU'' denotes the ground-truth IoU between the predicted box (dashed orange) and the ground-truth box (solid red). ``CLS'' denotes the classification score. In ``MAE/Ours: x(y)'', ``x'' denotes the calculated quality score using MAE/Ours metrics, and ``y'' denotes category-aware quality scores by combining ``CLS'' and ``x''.
	}
	\label{fig:MAE}
\end{figure}

Specifically, as shown in Figure~\ref{fig:MAE}, two sets of predicted boxes and their corresponding ground-truth (GT) boxes are displayed, with the solid red boxes representing the GT boxes and the dashed orange boxes representing the predicted boxes. Observing Figure~\ref{fig:MAE}(a) and Figure~\ref{fig:MAE}(b), it is evident that the predicted box in (a) exhibits higher localization quality than the one in (b) (i.e., $\text{GT IoU}_a = 0.88 > \text{GT IoU}_b = 0.49$).  
However, MAE fails to capture this distinction effectively ($\text{MAE: x}_a = 0.94$ vs. $\text{MAE: x}_b = 0.84$). Consequently, when the classification score (``CLS'' in Figure~\ref{fig:MAE}) does not align with the actual quality of the predicted box ( i.e., $\text{CLS}_a = 0.82 < \text{CLS}_b = 0.97$), the final category-aware MAE scores after incorporating CLS lead to an incorrect ranking ($\text{MAE: (y)}_a = 0.77 < \text{MAE: (y)}_b = 0.81$), contradicting the actual localization quality. In contrast, the proposed metric exhibits greater discriminative power($\text{Ours: x}_a = 0.83$ vs. $\text{Ours: x}_b = 0.62$) and maintains the correct ranking even when affected by inaccurate CLS ($\text{Ours: (y)}_a = 0.68 > \text{Ours: (y)}_b = 0.60$).

\subsection{A.2. Discussion on KLD-based Integration Metric}

To address the limitations of the MAE-based metric, we attempt to normalize both $H_i$ and $F_i$ into probability distributions $L$ and $G$, and introduce the widely-used KLD as the integration metric. KLD between the probability distributions $L$ and $G$ could be calculated as follows:
\begin{equation}
	\label{eq:kld}
	S(L, G) = \exp ^{-\text{KLD}(L, G)} = \exp^{-\sum_{p \in P(b_i)} L(p) \log \frac{L(p)}{G(p)}}.
\end{equation}

As illustrated in Figure 4 of the main manuscript, the KLD-based metric provides better distinction. However, KLD inherently assigns greater weight to pixels with higher heatmap values, which may conflict with the requirements of our application scenario. In our case, it is crucial to treat all pixels equally when assessing pixel-level spatial consistency. Consequently, this weighting bias can lead to an overestimation of the quality score in certain cases.

\begin{figure}[t]
	\centering
	\includegraphics[width=0.48\textwidth]{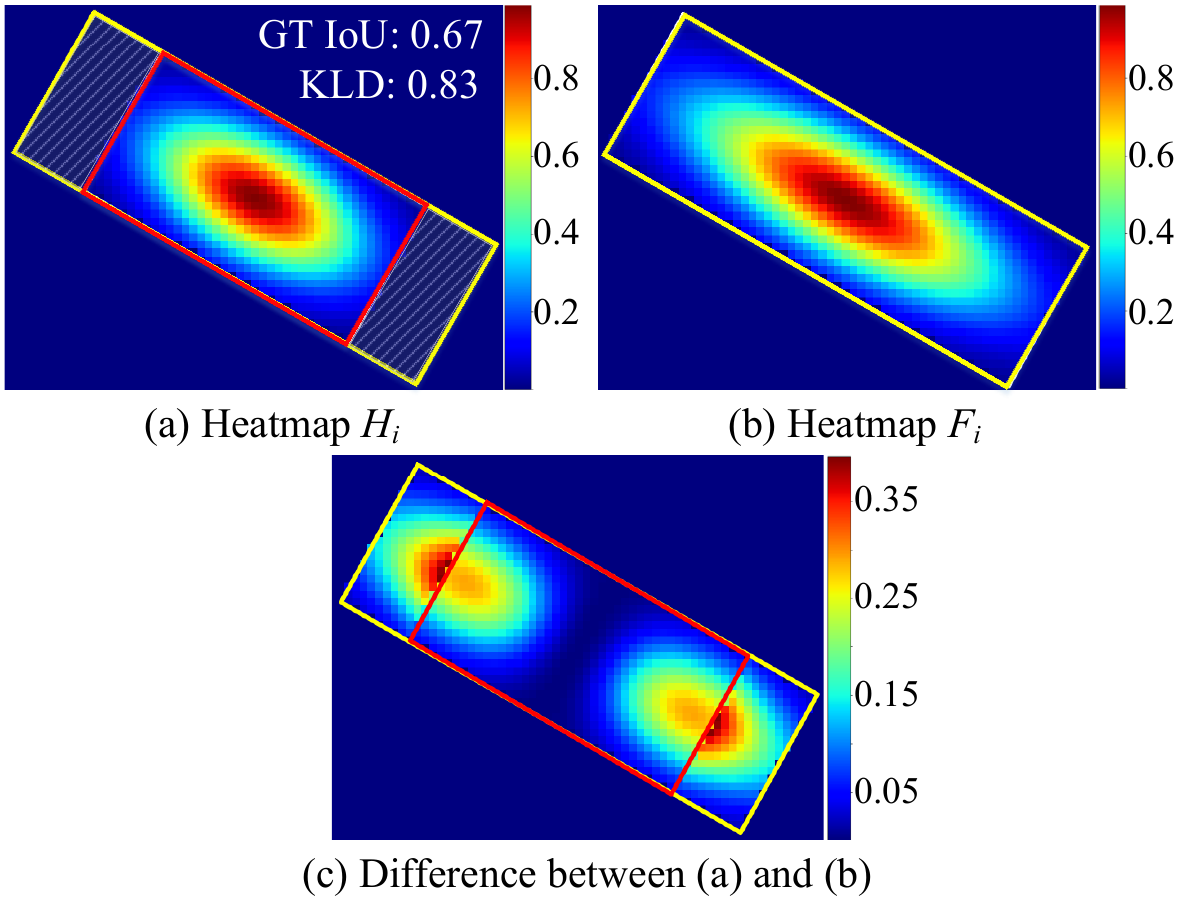}
	\caption{Illustrative cases demonstrating why KLD may conflict with the requirements of our application context. The predicted box is marked with yellow, and the GT box is marked with red. The non-overlapping regions between the predicted and GT boxes are highlighted with an upper diagonal pattern. ``GT IoU'' denotes the GT IoU between the predicted box (yellow) and the GT box (red). ``KLD'' denotes the calculated quality score using KLD.}
	\label{fig:KLD}
\end{figure}

Specifically, Figure~\ref{fig:KLD} illustrates a representative scenario where the predicted box (yellow) has an aspect ratio of 3:1, while the GT box (red) has an aspect ratio of 2:1. The non-overlapping regions between the predicted and GT boxes are highlighted with an upper diagonal pattern in Figure~\ref{fig:KLD}(a). 
According to Eq.~\eqref{eq:kld}, the calculated KLD values within these non-overlapping regions are zero, as the distribution $L(p)$ in these areas is also zero. As a result, the discrepancies between distributions  $L(p)$ and $G(p)$ in these regions are neglected. Notably, the differences in these areas are evidently greater than those in other areas, as shown in Figure~\ref{fig:KLD}(c). Consequently, the quality score calculated based on the KLD (\(\text{KLD} = 0.83\) in Figure~\ref{fig:KLD}(a)) is significantly higher than the actual GT IoU value (\(\text{GT IoU} = 0.67\)).

\section{B. Analysis and Implementation of PQA-Lite}

To reduce the computational overhead of the full PQA while preserving its performance benefits, we propose a lightweight variant, termed PQA-Lite. This section presents the implementation details of PQA-Lite and provides ablation studies on its different configurations.

\subsection{B.1. Implementation Details of PQA-Lite}

PQA-Lite improves inference efficiency by introducing two main modifications:

\begin{itemize}
	\item \textbf{Box Selection Strategy:} Instead of integrating pixel-level spatial consistency for all predicted boxes, we retain only the Top-$K$ boxes ranked by the probability that their centers are close to the GT box centers. This probability is derived from the estimated global position heatmap $H$, where each pixel's value (ranging from 0 to 1) indicates the likelihood of the pixel being near a GT box center. By sampling $H$ at the center of each predicted box, we obtain this probability. Well-localized boxes tend to have higher values, allowing us to filter out poorly localized predictions by selecting the top-$K$ boxes. The filtered boxes, which are excluded from spatial consistency integration, are assigned quality scores directly based on their sampled probabilities.
	
	\item \textbf{Pixel Sampling Strategy: } For each selected box, we reduce the number of sampled pixels used for spatial consistency integration. Let $N$ denote the number of pixels sampled in the original full PQA. PQA-Lite reduces this number to $\gamma N$, where $\gamma \in (0, 1)$ is a sampling ratio. A uniform subsampling is applied to the originally sampled pixels with the ratio of $\gamma$.
\end{itemize}

These modifications only affect the forward pass of the integration of pixel-level spatial consistency and do not alter the training pipeline. As a result, PQA-Lite can be implemented with minor changes to the inference logic and easily integrated into existing detection frameworks.

\begin{figure}[h]
	\centering
	\includegraphics[width=0.5\textwidth]{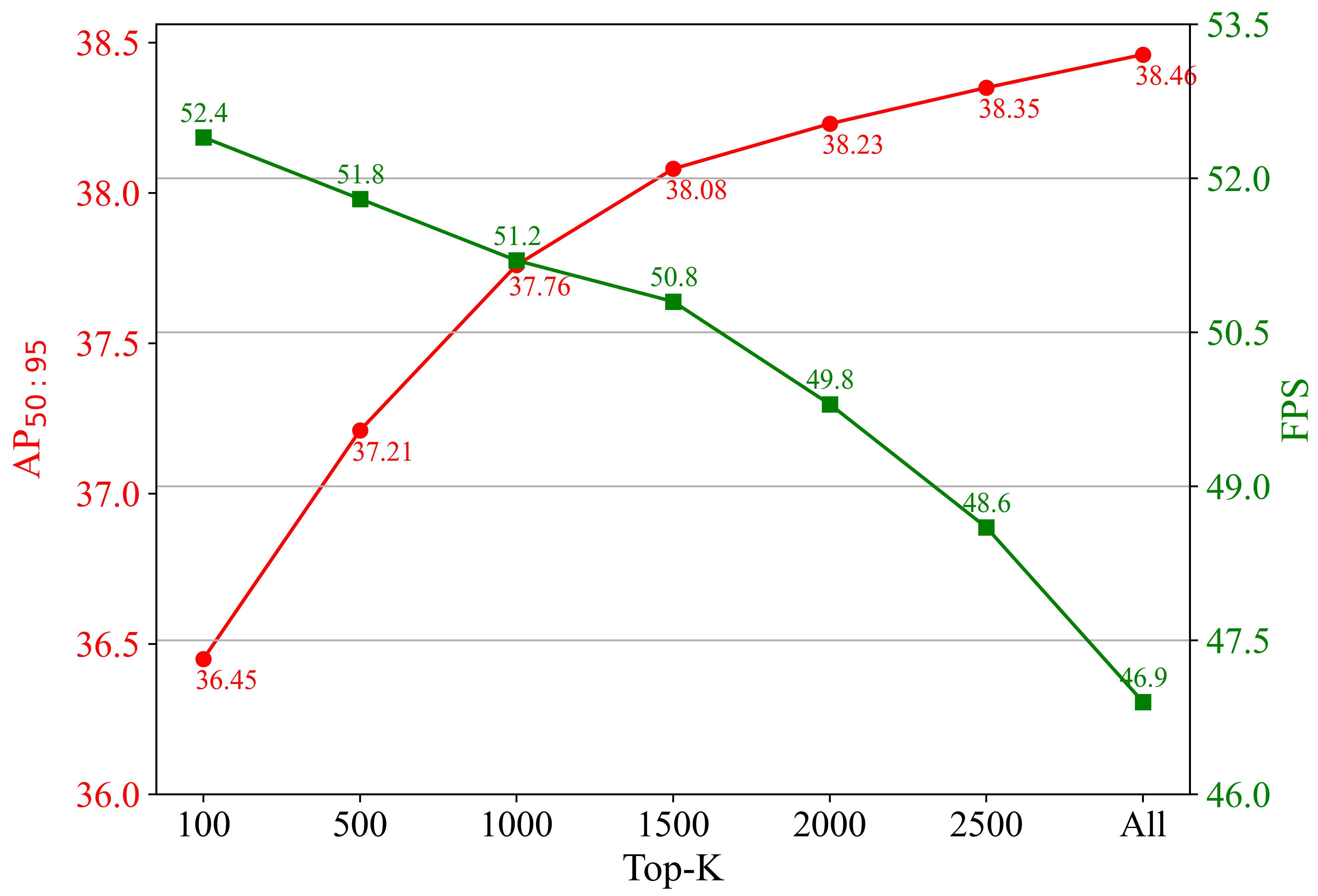}
	\caption{Impact of the Top-$K$ on AP$_{50:95}$ and FPS.}
	\label{fig:topk_curve}
\end{figure}

\begin{figure}[h]
	\centering
	\includegraphics[width=0.5\textwidth]{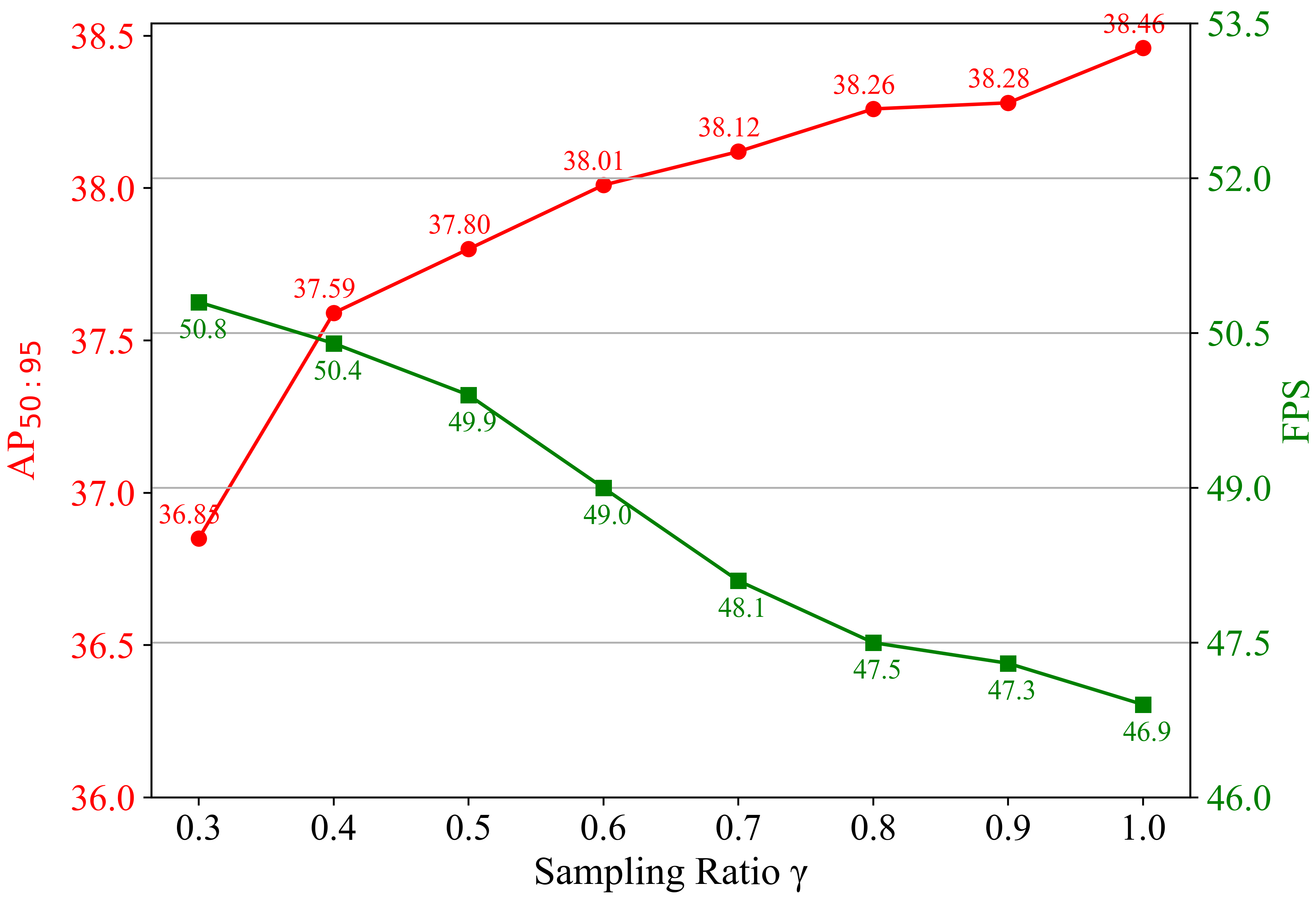}
	\caption{Impact of the sampling ratio $\gamma$ on AP$_{50:95}$ and FPS.}
	\label{fig:sample_curve}
\end{figure}

\subsection{B.2. Ablation Study on PQA-Lite}

\begin{table}[!t]
	\small
	\centering
	\renewcommand{\arraystretch}{1.2}
	\setlength{\tabcolsep}{3.0mm}{
		\begin{tabular}{c|ccccc}
			\hline
			$\lambda$ & 0.1 & 0.5 & 1.0 & 1.5 & 2.0 \\
			\hline
			\hline
			AP$_{50:95}$ & 36.16  & 37.78 & 38.05 & {\bf 38.46} & 38.07 \\
			AP$_{50}$   & 65.24 & 67.35 & 66.95 & {\bf 67.43} & 67.09 \\
			AP$_{75}$  & 34.70  & 35.47 & 35.84 & {\bf 36.44} & 35.76 \\
			\hline
			
	\end{tabular}}
	\caption{Results of PQA with various values of $\lambda$.}
	\label{tab_b5}
\end{table}

We conduct an ablation study to investigate how the top-$K$ selection and the sampling ratio $\gamma$ affect the trade-off between detection accuracy (AP$_{50:95}$) and inference speed (FPS). The experiments are carried out on the DOTA-v1.0 dataset, and all results are reported on the validation set. Specifically, we vary the number of selected boxes $K \in \{100, 500, 1000, 1500, 2000, 2500, \text{All}\}$ and the sampling ratio $\gamma \in \{0.3, 0.4, 0.5, 0.6, 0.7, 0.8, 0.9, 1.0\}$.

Figure~\ref{fig:topk_curve} and Figure~\ref{fig:sample_curve} illustrate the effects of the top-$K$ selection and the sampling ratio $\gamma$ on AP$_{50:95}$ and FPS, respectively. Compared to the full version of PQA ($K=\text{All}$, $\gamma=1.0$), which achieves an AP$_{50:95}$ of 38.46 and an inference speed of 46.9 FPS, selecting a subset (top-$1500$) of predicted boxes significantly boosts FPS to 50.8 while maintaining a competitive AP$_{50:95}$ of 38.08. Similarly, reducing the pixel sampling ratio $\gamma$ to $0.5$  per box improves FPS to 49.9 with a slight drop in AP$_{50:95}$ to 37.80. To achieve a favorable trade-off between detection accuracy and inference efficiency, we adopt $K=1500$ and $\gamma=0.5$ as the default configuration for PQA-Lite.

\section{C. Further Ablation Study on PQA}

This section presents additional analyses to better understand the behavior and reliability of the proposed PQA framework. We first examine the impact of the loss balancing parameter $\lambda$ on detection performance. Then, we conduct a controlled perturbation experiment to evaluate the robustness of the pixel-level integration strategy under prediction noise. 

{\bf Different values of $\lambda$.} 
The scaling factor $\lambda$ in Eq.~(10) of the main manuscript controls the contribution of the LD loss $\mathcal{L}_{ld}$ to the overall training loss. As reported in Table~\ref{tab_b5}, increasing $\lambda$ from 0.1 to 1.5 gradually improves performance, while a further increase to 2.0 leads to degradation. The underlying reason is that small $\lambda$ weakens the supervision on the global position heatmap, whereas large $\lambda$ causes it to interfere with localization learning. We therefore set $\lambda = 1.5$ in all experiments.

{\bf Robustness of pixel-level integration strategy.}
To assess the robustness of PQA's pixel-level integration strategy against prediction errors, we conduct a controlled perturbation experiment. Specifically, we randomly select a proportion $\delta_1$ of pixels in the predicted global position heatmap $H$ to perturb their values by a relative amount $\delta_2$ (within the range $[0,1]$). Detection accuracy is evaluated under varying $\delta_1$ and $\delta_2$ settings. For comparison, the same perturbation is applied to the box-level IoU prediction method by injecting noise into the predicted IoU values.
As shown in Table~\ref{tab:noise_robustness}, the box-level method exhibits a significant performance drop as perturbation increases; at $\delta_1 = 0.3$ and $\delta_2 = 0.4$, its AP$_{50:95}$ decreases by $1.21$. In contrast, PQA shows only a $0.42$ drop under the same conditions and maintains consistently stable performance across all perturbation levels. These results demonstrate the superior robustness of the proposed pixel-level integration strategy over conventional box-level IoU prediction.

\begin{table}[!t]
	\scriptsize
	\centering
	\renewcommand{\arraystretch}{1.2}
	\setlength{\tabcolsep}{1.7mm}{
		\begin{tabular}{c|c|ccc|ccc}
			\hline
			\multirow{2}{*}{$\delta_1$} & \multirow{2}{*}{$\delta_2$} & \multicolumn{3}{c|}{Box-level IoU Prediction} & \multicolumn{3}{c}{Pixel-level PQA} \\
			\cline{3-8}
			& & AP$_{50:95}$ & AP$_{50}$ & AP$_{75}$  & AP$_{50:95}$ & AP$_{50}$ & AP$_{75}$  \\
			\hline
			\hline
			0.0 & 0.0 & 36.79 & 66.52 & 35.12 & 38.46 & 67.43 & 36.44 \\
			0.1 & 0.2 & 36.40$_{{(\downarrow 0.39)}}$ & 66.11 & 34.40 & 38.45$_{{(\downarrow 0.01)}}$ & 67.43 & 36.28 \\
			0.2 & 0.3 & 35.80$_{{(\downarrow 0.99)}}$ & 65.48 & 33.58 & 38.27$_{{(\downarrow 0.19)}}$ & 67.27 & 36.18\\ 
			0.3 & 0.4 & 35.58$_{{(\downarrow 1.21)}}$ & 65.33 & 33.40 & 38.02$_{{(\downarrow 0.42)}}$ & 67.17 & 36.15 \\
			\hline
	\end{tabular}}
	\caption{Detection performance under different perturbation levels.}
	\label{tab:noise_robustness}
\end{table}

\end{document}